\documentclass[lettersize,journal]{IEEEtran}
\usepackage{amsmath,amsfonts}
\usepackage{algorithmic}
\usepackage{algorithm}
\usepackage{pifont}
\usepackage{array}
\usepackage[caption=false,font=normalsize,labelfont=sf,textfont=sf]{subfig}
\usepackage{textcomp}
\usepackage[table,xcdraw]{xcolor}
\usepackage{colortbl}
\usepackage{stfloats}
\usepackage{url}
\usepackage{bbding}
\usepackage{booktabs}
\usepackage{diagbox}
\usepackage{multirow}
\usepackage{verbatim}
\usepackage{amsmath}
\usepackage{color}
\usepackage{graphicx}
\usepackage{cite}
\usepackage{soul}
\usepackage{makecell}
\usepackage{hyperref}
\hypersetup{
colorlinks=true,
linkcolor=black,
citecolor=black
}
\begin{document}

\title{RSDehamba: Lightweight Vision Mamba for Remote Sensing Satellite Image Dehazing}

\author{Huiling Zhou, Xianhao Wu, Hongming Chen, Xiang Chen, Xin He
\thanks{Co-corresponding author: Xiang Chen and Xin He}

\thanks{Huiling Zhou, Xianhao Wu and Hongming Chen are with the College of Electronic Information Engineering, Shenyang Aerospace University, Shenyang 110135, China.}

\thanks{Xiang Chen is with the School of Computer Science and Engineering, Nanjing University of Science and Technology, Nanjing 210094, China (e-mail: chenxiang@njust.edu.cn).}

\thanks{Xin He is with the School of Basic Sciences for Aviation, Naval Aviation University, Yantai 264001, China.}
}

\maketitle

\begin{abstract}
Remote sensing image dehazing (RSID) aims to remove nonuniform and physically irregular haze factors for high-quality image restoration.
The emergence of CNNs and Transformers has taken extraordinary strides in the RSID arena.
However, these methods often struggle to demonstrate the balance of adequate long-range dependency modeling and maintaining computational efficiency.
To this end, we propose the first a lightweight network on the mamba-based model called RSDhamba in the field of RSID. Greatly inspired by the recent rise of Selective State Space Model (SSM) for its superior performance in modeling linear complexity and remote dependencies, our designed RSDehamba integrates the SSM framework into the U-Net architecture.
Specifically, we propose the Vision Dehamba Block (VDB) as the core component of the overall network, which utilizes the linear complexity of SSM to achieve the capability of global context encoding.
Simultaneously, the Direction-aware Scan Module (DSM) is designed to dynamically aggregate feature exchanges over different directional domains to effectively enhance the flexibility of sensing the spatially varying distribution of haze.
In this way, our RSDhamba fully demonstrates the superiority  of spatial distance capture dependencies and channel information exchange for better extraction of haze features.
Extensive experimental results on widely used benchmarks validate the surpassing performance of our RSDehamba against existing state-of-the-art methods.
\end{abstract}


\section{Introduction}
\IEEEPARstart{I}{mage} 
dehazing is typically utilized in the process of remote sensing (RS) image acquisition. 
As an atmospheric disturbance factor, haze tends to degrade the visual quality of RS images, thus limiting their usability in the fields of geological survey, environmental monitoring, urban development planning, $\emph{etc}$.
Hence, constructing a high-performance Remote Sensing Image Dehazing (RSID) architecture to recover high-quality remote sensing images has become a hot issue in the current research \cite{cong2020discrete,gui2021comprehensive,chen2022unpaired,chen2021hybrid}.

\begin{figure}[!t]
	\centering 	
\includegraphics[width=\columnwidth]{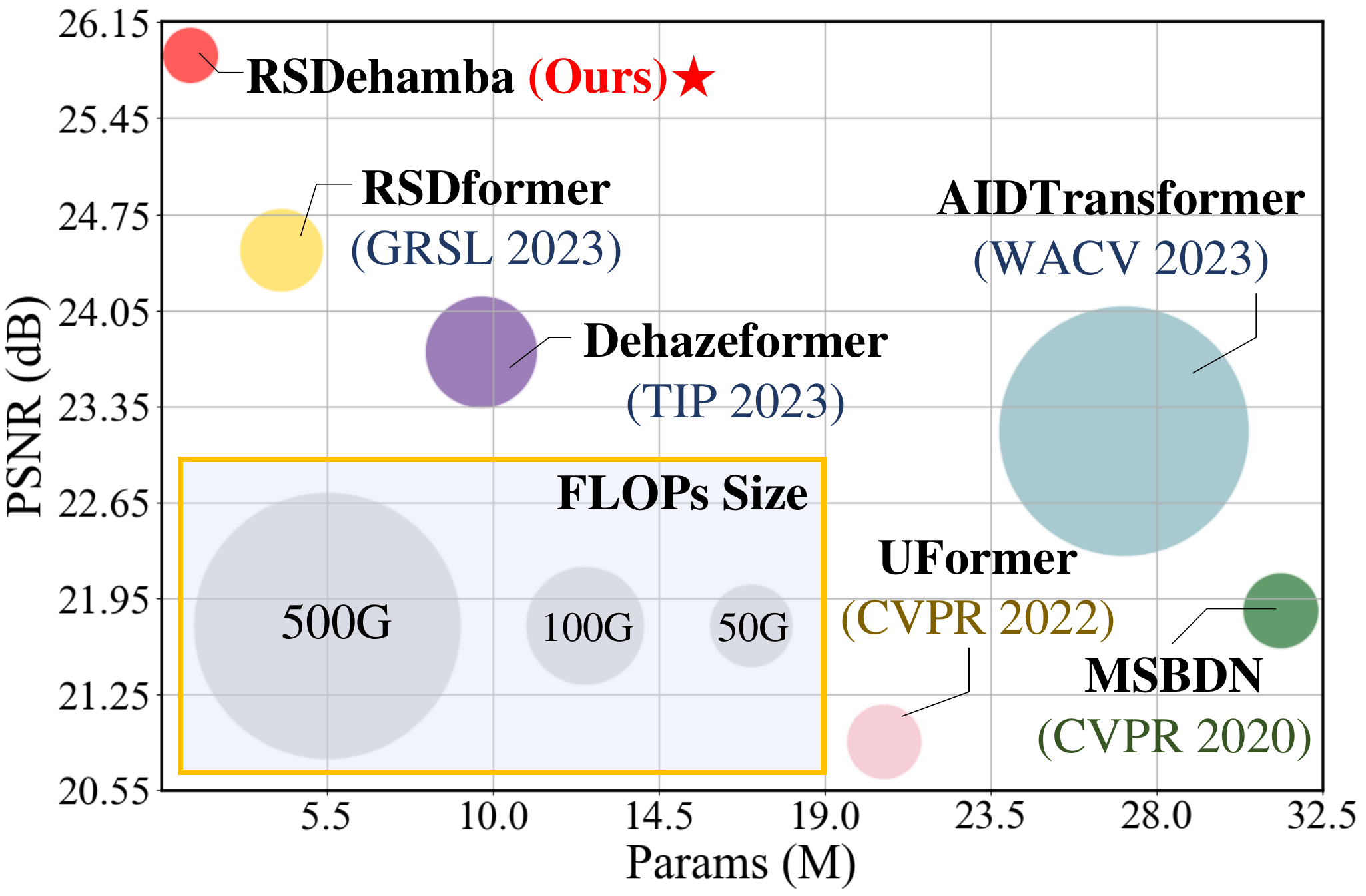}
\vspace{-7mm}
    \caption{Model complexity and performance comparisons of the proposed method and other state-of-the-art models on the Sate1K dataset in terms of PSNR, model parameters and FLOPs. The area of each circle denotes the number of FLOPs. Here, FLOP calculation is based on image sizes of $256 \times 256$. Our RSDhamba achieves the SOTA performance.}
	\label{fig1}
 	\vspace{-3mm}
	 \end{figure}

To compensate for this urgent demand, early RS image dehazing methods explored by researchers originate from manual priors on image statistical features, such as dark channel priors\cite{he2010single} and virtual point clouds\cite{long2013single}.
Despite their simplicity and low computational complexity, these traditional methods usually perform well only under specific conditions, making it difficult to accurately capture clear features under complex nonuniform haze conditions.
Driven by the success of deep learning on high-level visual tasks, many approaches to image dehazing have favored more efficient CNN architectures such as AOD-Net\cite{li2017aod}, FFA-Net\cite{qin2020ffa}, MSBDN\cite{dong2020multi}, FCTF-Net\cite{li2020coarse}, and so on.
Although the convolutional operation effectively models local connectivity, its limited local receptive domain, which severely hampers the ability to perform long-range dependent features, weakens the performance of CNN-based methods.

Motivated by the global dependency of the captured content, Transformers\cite{guo2022image,song2023vision,kulkarni2023aerial,song2023learning} are integrated into the field of RS image dehazing and show excellent performance. 
For instance, Dehamer\cite{guo2022image} implements RSID tasks by combining the feature strengths of CNNs and Transformers.
Dehazeformer\cite{song2023vision} proposes a multi-improved image dehazing Transformer network.
AIDTransformer\cite{kulkarni2023aerial} exploits a Transformer architecture with deformable attention.
However, RSID mostly deals with high-resolution images \cite{he2024learning}, and the self-attention mechanism in Transformers makes the secondary time complexity high, which  imposes a large computational burden.

\begin{figure*}[t]
	\centering
	\includegraphics[width=1.0\textwidth]{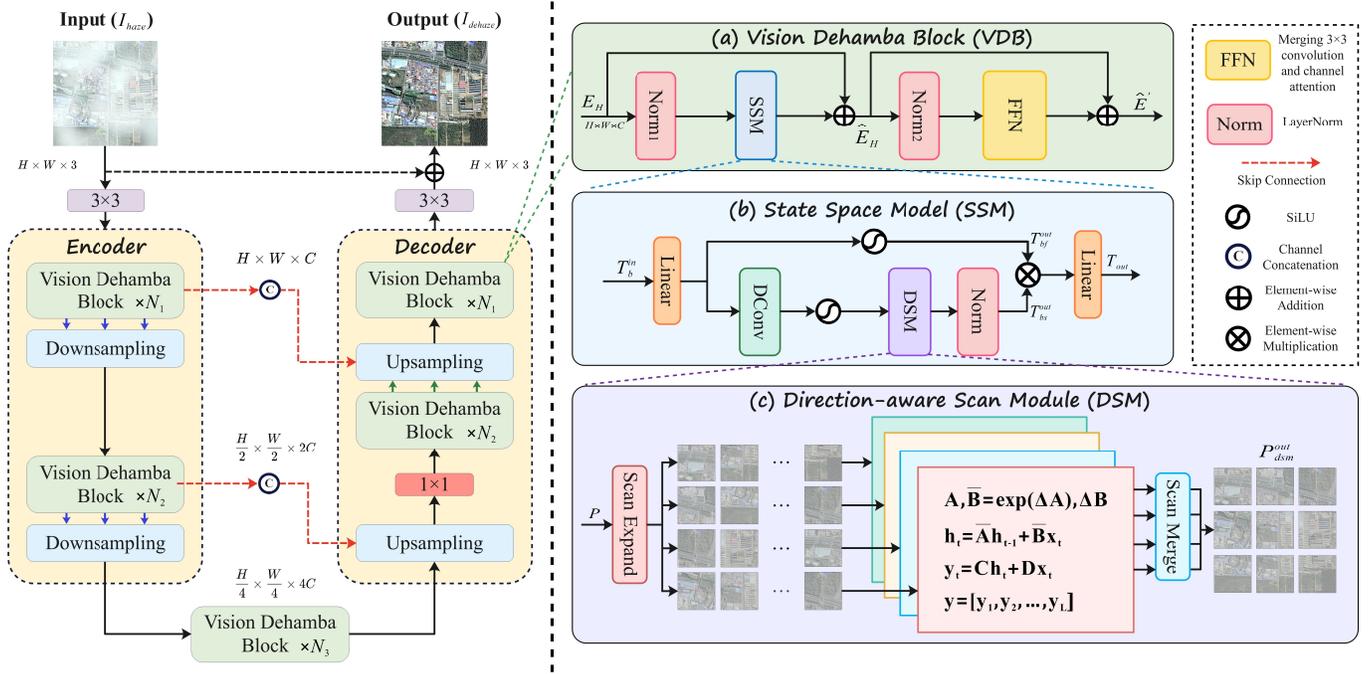}
	\caption{Overall architecture of RSDhamba. (a) RSDhamba consists of multiple Vision Dehamba Blocks (VDBs) using the U-Net architecture for image defogging tasks. (b) The VDB mainly consists of the State Space Model (SSM): a module for capturing local and global features.
(c) An important component of SSM: the Direction-aware Scan Module (DSM) performs perceptual modeling by scanning from four directions.}
	\label{fig2}
	\vspace{-3mm}
\end{figure*}

Recent advances have freshly introduced State Space Models (SSMs), especially the Mamba model\cite{gu2023mamba}, which captures global contextual features as well as linearly lower complexity for input tokens, exceeding the performance of both CNN-based and Transformer-based models.
Very recently, a handful of Mamba variants presented exceptional results in computerized high-level vision. MambaIR\cite{guo2024mambair} explores for the first time the role of state-of-the-art Mamba in image restoration. U-Mamba\cite{ma2024u} devises hybrid CNN-SSM blocks to build a generalized biomedical image segmentation network. 
However, Mamba remains underexplored for RSID, which motivates us to investigate vision Mamba to enhance the performance of RSID in remote modeling capabilities.

In this letter, we construct a lightweight mamba architecture in the RSID domain for the first time, called \textbf{RSDhamba}.
Specifically, the overall architecture of the model follows a three-level encoder-decoder network structure, with a Vision Dehamba Block (VDB) forming the backbone element of the network, and integrates the SSM framework into the U-Net architecture.
To better capture the nonuniform features of haze in spatial distribution, we introduce the Direction-aware Scan Module (DSM) to perceive the spatially varying distribution of haze from different directions and realize the dynamic feature fusion in multiple directions.
With the above design, we construct the RSDhamba model with a strong capability of capturing the far-proximity binding dependencies while maintaining the computational efficiency, which enhances the recovery performance of RSID, thus significantly improving the quality and accuracy of the recovered images.

The main contributions of our work are as follows:
\begin{itemize}
 \item {We propose the first lightweight RSDhamba network architecture on the Mamba-based model to fill the gap of RSID performance enhancement, incorporating a state-space model (SSM) to enrich the global context of U-Net for image restoration.}
\item {We design the Direction-aware Scan Module (DSM), which senses the spatially varying haze distribution during the defogging process through dynamic scanning, aiming to enhance the transfer of effective information between different regions.}
\item {Our experiments show that RSDhamba achieves extremely favorable performance on a variety of RSID datasets, outperforming current SOTA methods while maintaining low computational complexity.}
\end{itemize}

\section{PROPOSED METHOD}
\label{sec2}
In this section, we first present the overall network architecture of the proposed RSDehamba. Afterwards,We introduce the benchmark module of our model, Vision Dehamba Block (VDB), which is based on the core State Space Module (SSM) mechanism and effectively utilizes Mamba for remote modeling capabilities and high-dimensional processing of visual data. Finally, we detail the Direction-aware Scanning Module (DSM) to better capture the long-term dependence of the haze image information flow in four directions for each sequence, addressing the limitations of Mamba one-way modeling.
\vspace{-3mm}
\subsection{Network Architecture}
To ensure effective feature fusion of haze images, we adopted a multi-scale UNet architecture based on the Vision Dehamba Block (VDB) as the network backbone. Specifically, the model architecture is illustrated in Fig. \ref{fig2}.

Given a degraded haze image \( I_{haze} \in \mathbb{R}^{H \times W \times 3} \), shallow feature embedding \( E_{0} \in \mathbb{R}^{H \times W \times C} \) is obtained first through a $3\times3$  convolution. Then, the projected features pass through a three-level symmetric encoder-decoder UNet architecture. Each level consists of multiple VDB, with the number of blocks varying with the level, ranging from [2, 3, 3]. These blocks are applied to progressively learn and recover clear details of remote sensing images. The decoder takes low-resolution feature \(E_{l} \in \mathbb{R}^{\frac{H}{4} \times \frac{W}{4} \times 4C} \) as input and achieves cross-scale information aggregation through skip connections. After cascaded the $1\times1$ convolution and two upsampling operations, \(E_{l} \) is transformed into deep features \( E_{d}\in\mathbb{R}^{H \times W \times 2C} \). Finally, refinement of features \( E_{d} \) is performed using a $3\times3$ convolution to generate residual image \( R_{i}\in\mathbb{R}^{H \times W \times 3} \). The final reconstruction result is obtained by: $I_{dehaze}=$ $\mathcal{F}(I_{haze})+I_{haze},$  where $\mathcal{F}(\cdot)$ is the overall network, trained by minimizing the following loss function: 
\vspace{-1mm}
\begin{equation}
\mathcal{L}=\left\|I_{dehaze}-I_{gt}\right\|_1,
\label{eq:L1}
\end{equation}
where $I_{gt}$ represents the ground truth image, $\|\cdot\|_{1}$ represents the $L_{1}\mathrm{-norm}$.
By integrating the U-Net architecture design, RSDehamba can more effectively capture rich global frequency information and fusion of features at different scales. Next, we will elaborate on the details of the three key design elements mentioned in this letter.

\begin{figure}[!t]
	\centering 	
\includegraphics[width=\columnwidth]{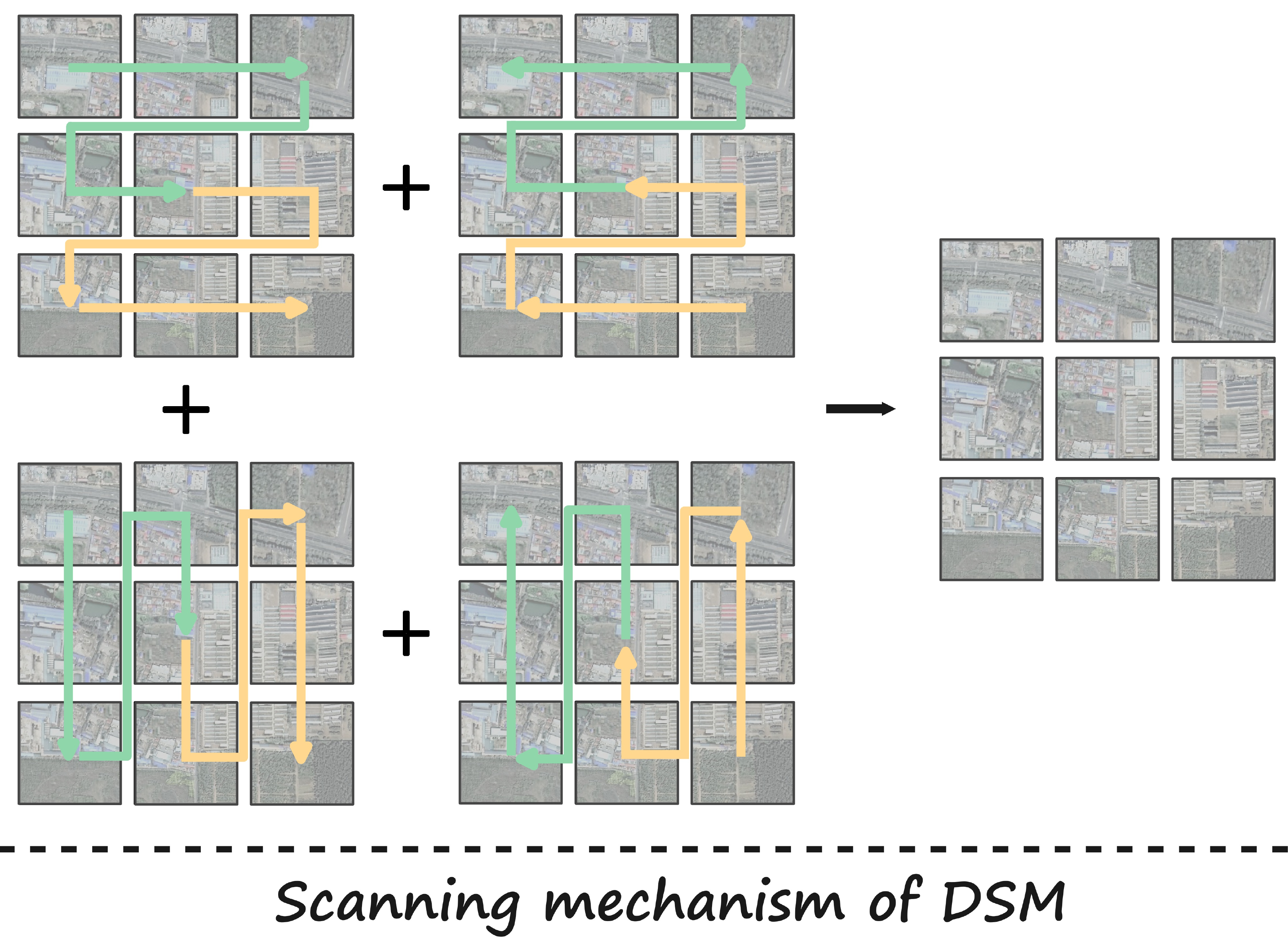}
\vspace{-7mm}
    \caption{The scanning mechanism of DSM is the fusion of regional haze features from four different directions to better perceive the distribution of spatial changes in haze.}
	\label{fig3}
 	\vspace{-3mm}
	 \end{figure}
\vspace{-3mm}  
\subsection{Vision Dehamba Block}
Haze images typically exhibit nonuniform distribution and irregular physical forms. Therefore, capturing features from both local and non-local regions is equally crucial in RS dehazing models. To address this problem, we design the Vision Dehamba Block (VDB), leveraging an enhanced SSM mechanism to assist the vanilla Mamba in long-range spatial modeling. Specifically, the input undergoes a $3\times3$ convolution to generate shallow features \(E_H \in \mathbb{R}^{H \times W \times C}\) at the first stage, then using the layer normalization for spatial data long-term dependencies by the State Space Module (SSM). Finally, a residual connection is applied to obtain the output $\hat{E}_{H}$.

To enhance the interaction capability of spatial feature modeling in foggy images, we incorporate another normalization layer in the second stage. Further, by utilizing a $3\times3$ convolution, we aim to improve feature extraction efficiency and reduce channel redundancy. Ultimately, the features are passed through a FFN layer for deep feature extraction, and the resulting output is then combined with the residual. This process is defined as follows:
\begin{equation}
\hat{E}_{H} = E_H + \text{SSM}(\text{Norm}_1(E_H)),
\label{eq:VDB1}
\end{equation}
\vspace{-4mm}
\begin{equation}
\hat{E}^{\prime} =\hat{E}_{H} + \text{FFN}(\text{Norm}_2(\hat{E}_{H})),
\label{eq:VDB2}
\end{equation}
where \(E_H\) represents the shallow feature embedding obtained from the input, and \(\text{Norm}_1\) denotes the first normalization layer applied to the input feature. The final output of the first stage is denoted as \(\hat{E}_{H}\). Additionally, \(\text{Norm}_2\) represents the second normalization layer applied to the input features, and \(\hat{E}^{\prime}\) denotes the final output of the proposed VDB.

\begin{table*}[t]
	\caption{Comparison of quantitative results on SateHaze 1k datasets. \textbf{Bold} and \underline{underline} indicate the best and second-best results.}
	\vspace{-3mm}
	\resizebox{1\textwidth}{!}{
	\begin{tabular}{cc|cc|cc|cc|cc|cc}
	\Xhline{1pt}
	\multicolumn{2}{c|}{Benchmark Datasets}                                                   & \multicolumn{2}{c|}{Thin Haze}    & \multicolumn{2}{c|}{Moderate Haze}    & \multicolumn{2}{c|}{Thick Haze}   & \multicolumn{2}{c|}{Average}    & \multicolumn{2}{c}{Complexity} \\ \hline
	\multicolumn{2}{c|}{Metrics}                                                    & PSNR           & SSIM            & PSNR           & SSIM            & PSNR           & SSIM            & PSNR           & SSIM           & \multicolumn{2}{c}{{\#Params}}            \\ \hline
	\multicolumn{1}{c|}{\multirow{1}{*}{Prior-based methods}}       & DCP \cite{he2010single}            & 13.45          & 0.6977          & 9.78          & 0.5739          & 10.90          & 0.5715          & 11.38          & 0.6144     & \multicolumn{2}{c}{{-}}     \\ \hline
	\multicolumn{1}{c|}{\multirow{8}{*}{CNN-based methods}}         & AOD-Net \cite{li2017aod}          & 18.74          & 0.8584          & 17.69          & 0.7969          & 13.42          & 0.6523          & 16.60          & 0.7692          & \multicolumn{2}{c}{{0.02M}}          \\
	\multicolumn{1}{c|}{}                                           & LD-Net \cite{ullah2021light}       & 17.83          & 0.8568          & 19.80          & 0.8980          & 16.60          & 0.7649          & 18.08          & 0.8399          & \multicolumn{2}{c}{{0.03M}}          \\
	\multicolumn{1}{c|}{}                                           & GCANet \cite{chen2019gated}       & 22.27          & 0.9030          & 24.89          & 0.9327          & 20.51          & 0.8307          & 22.56         & 0.8888         & \multicolumn{2}{c}{{0.70M}}          \\
	\multicolumn{1}{c|}{}                                           & GridDehazeNet \cite{liu2019griddehazenet}       & 20.04          & 0.8651          & 20.96          & 0.8988          & 18.67          & 0.7944          & 19.89          & 0.8528          & \multicolumn{2}{c}{{0.96M}}          \\
	\multicolumn{1}{c|}{}                                           & FFA-Net \cite{qin2020ffa}       & 22.30          & 0.9072          & 25.46          & 0.9372          & 20.84          & 0.8451          & 22.87          & 0.8965          & \multicolumn{2}{c}{{4.68M}}          \\
	\multicolumn{1}{c|}{}                                           & MSBDN \cite{dong2020multi}      & 18.02          & 0.7029          & 20.76          & 0.7613          & 16.78          & 0.5389          & 18.52         & 0.6677         & \multicolumn{2}{c}{{31.35M}}                \\
	\multicolumn{1}{c|}{}                                           & FCTF-Net \cite{li2020coarse}       & 20.06          & 0.8808          & 23.43          & 0.9265         & 18.68          & 0.8020          & 20.72          & 0.8698          & \multicolumn{2}{c}{{0.16M}}            \\
	\multicolumn{1}{c|}{}                                           & M2SCN \cite{li2022m2scn}       & \underline{25.21}          & 0.9175          & 26.11          & 0.9416          & 21.33          & 0.8289          & 24.22          & 0.8960          & \multicolumn{2}{c}{{-}}          \\ \hline
	\multicolumn{1}{c|}{\multirow{6}{*}{Transformer-based methods}} & Restormer \cite{zamir2022restormer}    & 24.97          & \underline{0.9248}          & \underline{26.77}          & \underline{0.9452}          & 21.28          & 0.8362          & \underline{24.34}          & 0.9021          & \multicolumn{2}{c}{{26.13M}}           \\
	\multicolumn{1}{c|}{}                                           & Dehamer \cite{guo2022image}    & 20.94          & 0.8717           & 22.89          & 0.8708          & 19.80          & 0.8086          & 21.21          & 0.8504          & \multicolumn{2}{c}{{132.45M}}          \\
	\multicolumn{1}{c|}{}                                           & DehazeFormer \cite{song2023vision}          & 23.92          & 0.9121          & 25.94         & 0.9445          & 22.03          & 0.8373          & 23.96          & 0.8980         & \multicolumn{2}{c}{{4.64M}}           \\
	\multicolumn{1}{c|}{}                                           & AIDTransformer  \cite{kulkarni2023aerial} & 23.12 &0.9052 & 25.08 & 0.9124 & 20.56 & 0.8325 & 22.92 & 0.8834 & \multicolumn{2}{c}{{20.33M}}  \\ 
     \multicolumn{1}{c|}{}                                           & RSDformer \cite{song2023learning}          & 24.06          & 0.9177         & 25.97         & 0.9390          & \underline{22.87}          & \underline{0.8646}          & 24.30          & \underline{0.9071}        & \multicolumn{2}{c}{{5.35M}}           \\ \hline
        \multicolumn{1}{c|}{\multirow{1}{*}{Mamba-based method}} & RSDehamba     &\textbf{26.75}           & \textbf{0.9306}          & \textbf{27.45}          & \textbf{0.9468}         & \textbf{23.53}          & \textbf{0.8698}         &  \textbf{25.91}        & \textbf{0.9157}          & \multicolumn{2}{c}{{1.80M}}           \\  \Xhline{1pt}
\end{tabular}
}
\vspace{-5mm}
	\label{table1}	
\end{table*}

\begin{figure*}[t]
	\centering
	\includegraphics[width=1.0\textwidth]{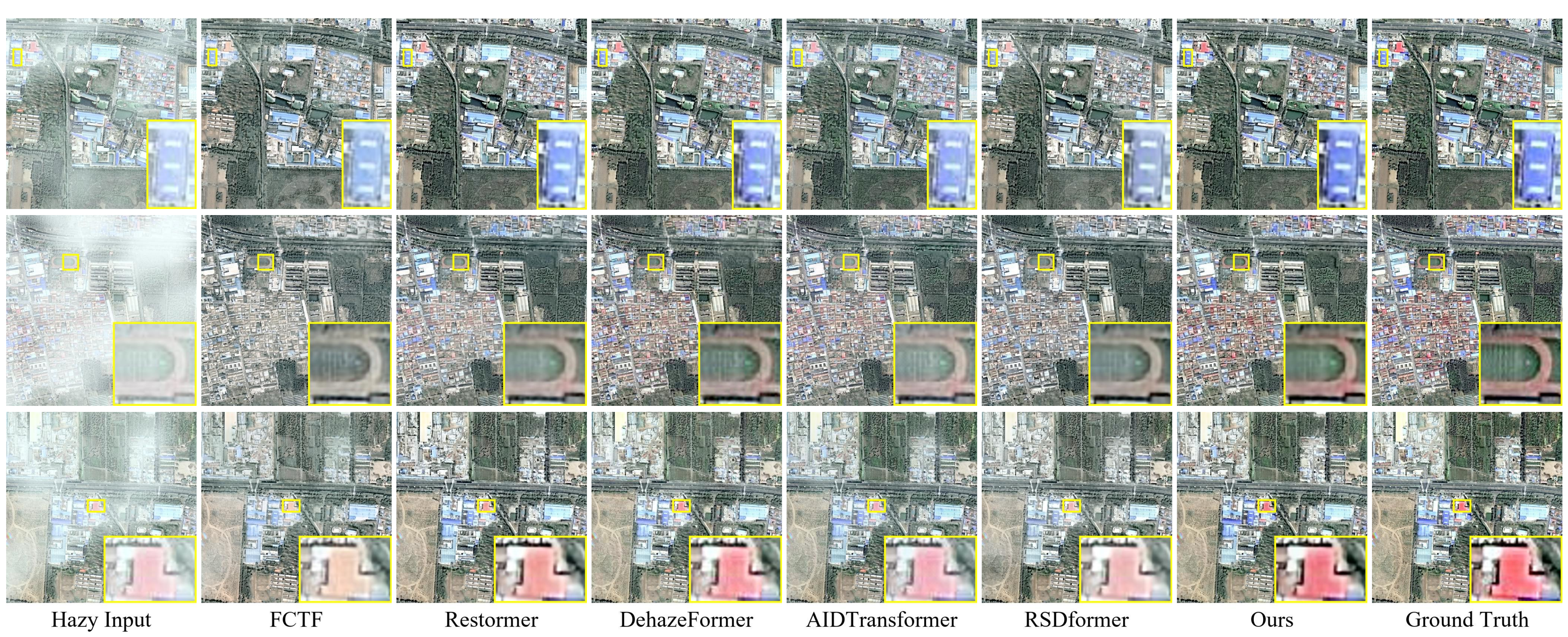}
	\vspace{-7mm}
	\caption{Comparison of visual results for the SateHaze1k dataset, with colored boxes highlighting obvious differences. Zoom in for the best view.}
	\label{fig4}
	\vspace{-3mm}
\end{figure*}
\vspace{-3mm}
\subsection{State Space Model}
To finely capture the complex structures and textures in haze images, we designed the State Space Model (SSM) to learn long-term spatial dependencies in the images.
Structured state space sequence models (SSMs) \cite{guo2024mambair} encode a 1D sequential input \( x(t) \in \mathbb{R} \) into an output \( y(t) \in \mathbb{R} \) via an implicit latent state \( h(t) \in \mathbb{R}^N \). These models are characterized by four parameters (\(\Delta\), \(A\), \(B\), \(C\)) as follows:
\vspace{-1mm}
\begin{equation}
{h_t=\overline{A}}{h_{t-1}+\overline{B}}{x_t}.
\label{eq:SSM1}
\end{equation}
\vspace{-5mm}
\begin{equation}
{y_t=Ch_t+Dx_t}. 
\label{eq:SSM2}
\end{equation}
\vspace{-5mm}

Eventually discretized into sequences in SSM blocks as ${y=[y_1,y_2,...,y_L]}$.
Overall, the State Space Model (SSM) is a dual-branch component.
In the first branch, the input simply passes through a linear layer and a SiLU activation function. 
Simultaneously, the input sequentially goes through a linear layer, depth-wise separable convolution, and a SiLU activation function in the second branch, followed by feeding into the DSM to perform global scanning and effective fusion of feature information between different haze regions.
Finally, it is processed through the normalization layer as the second branch output.
Considering the interaction of haze features between different branches, we merge the output features of the two branches at the end of the network and then obtain the output through a linear layer, effectively enhancing the dehazing performance and generalization ability of the model.
\vspace{-3mm}
\begin{equation}
\mathbf{T}_{bf}^{out}=\phi(Linear(\mathbf{T}_{b}^{in})), 
\label{eq:SSM1}
\end{equation}
\vspace{-4mm}
\begin{equation}
\mathbf{T}_{bs}^{ou\boldsymbol{t}}=\mathrm{Norm}\{DSM\left(\phi \left(\mathrm{DConv}\left({Linear}(\mathbf{T}_{b}^{in})\right)\right)\right)\}, 
\label{eq:SSM2}
\end{equation}
\vspace{-4mm}
\begin{equation}
\mathbf{T}_{out}={Linear}(\mathbf{T}_{bf}^{out}\odot\mathbf{T}_{bs}^{out}),
\label{eq:SSM3}
\end{equation}
where $\mathbf{T}_{b}^{in}$ represents the feature input of the SSM model, $ T_{\text{bf}}^{\text{out}}$ and $T_{\text{bs}}^{\text{out}}$ represent the outputs of the first and second branches, respectively.$\phi(\cdot)$ denotes the SiLU activation function, $ \text{DConv}(\cdot) $ stands for depth-wise separable convolution, $\odot$ indicates the Hadamard product.
\vspace{-4mm}
\subsection{Direction-aware Scan Module}
Unlike Transformer's self-attention mechanism \cite{chen2023learning}, which operates in a two-dimensional plane, Visual mamba\cite{gu2023mamba} has recently integrated image modeling to solve the quadratic complexity of its input sequence. It is worth noting that the standard mamba is designed for the feature sequence of one-dimensional haze images. To enhance Mamba's multidimensional modeling ability in RS image dehazing, we introduced Direction-aware Scan Module (DSM) as shown in Fig. \ref{fig3}. Firstly, we use the two-dimensional plane feature  \( P \)$\in $\( \mathbb{R}^{B \times C \times H \times W} \) as the input of the scanning module, and flatten it into a one-dimensional sequence $\mathbf{Q}_{seq}^{l}$$\{l=0,1,2,3\}$ of length l, conducting a forward and backward scanning of four paths from top left to bottom right and from bottom right to top left. The first scanning process can be defined as:
\vspace{-1mm}
\begin{equation}
\mathbf{Q}_{seq}^{l0}=[\mathbf{u}_{lu}^{1};\mathbf{u}_{uf}^{2};\cdots;\mathbf{u}_{ld}^{l-n};\mathbf{u}_{df}^{l-n+1};\cdots;\mathbf{u}_{rd}^{\mathbf{L}}], \\
\label{eq:DSM1}
\end{equation}
where \( u_{lu}^i \) represents the i-th token of the top-left scan, \( u_{uf}^j \) represents the j-th token of the upward forward scan, \( u_{ld}^{(l-n)} \) represents the (l-n)-th token of the left-downward scan, \( u_{df}^{(l-n+1)} \) represents the (l-n+1)-th token of the downward backward scan, and \( u_{rd}^L \) represents the last scan token of the right downward. Sending these token sequences \( Q_{\text{seq}}^l \) to the Mamba block captures the features for long-range modeling, resulting in \( P_{\tilde{s}} \) $\in $ \( \mathbb{R}^{B \times 4C \times HW} \). The sequences from all four directions are merged to obtain the globally fused feature output \( P_{\text{dsm}}^{\text{out}} \) $\in $ \( \mathbb{R}^{B \times C \times H \times W} \).

\section{ EXPERIMENT RESULTS}
In this section, we provide details of the experimental setup and implementation information. To accurately evaluate the proposed RSDehamba component, we perform a comprehensive image dehazing experiment on the SateHaze1K dataset with other baselines. Finally, we conduct ablation studies to confirm the effectiveness of RSDehamba.

\subsection{Implementation Details}
The block numbers $\{N_{1},N_{2},N_{3}\}$ in the model are $\{2,3,3\}$. During the training process, we used an Adam optimizer with patch size $256\times256$. The initial learning rate was set at 2 × $10^{-4}$, adaptive adjustment was made using the cosine annealing strategy, and gradually reduced to 1 × $10^{-6}$. The entire model was implemented using an NVIDIA RTX 4090 GPU on the PyTorch platform.
\vspace{-3mm}
\subsection{Qualitative Comparison}
For fair comparison, we conduct extensive experiments on the SateHaze1K benchmark. Consistent with previous work, we compute the PSNR and SSIM of the RGB channels as evaluation metrics. The results of the quantitative comparison on SateHaze1K are shown in Table \ref{table1}. Compared with the previous CNNs and Transformers, the model proposed in this paper makes very significant progress in both PSNR and SSIM. For example, in the heavy haze dataset, our method outperforms baseline methods such as DehazeFormer\cite{song2023vision} and RSDformer\cite{song2023learning} by more than 0.72dB. This shows that our method can handle different types of haze distributions more efficiently and accurately. As performing a quantitative comparison in terms of both trainable parameters and floating point operations (FLOPs) in Fig. \ref{fig1}, our model is extremely lightweight compared to other methods with only 1.80M.
\vspace{-4mm}
\subsection{Quantitative Comparison}
To verify the effectiveness of our proposed method in more depth, we visually compare it with other competing methods in Fig. \ref{fig4}. By amplifying the details, we can observe that RSDformer \cite{song2023learning} and DehazeFormer\cite{song2023vision} have difficulties in dealing with recovering image details. In contrast, RSDehamba efficiently learn and exchange global as well as local information, and thus it is able to exhibit superior dehazing performance in image details and retains clearer texture structures.
\begin{table}[t]\small
    \centering
    \caption{ABLATION ANALYSIS OF THE PROPOSED SSM CONSTITUENT ELEMENTS WHTHIN THE MODEL.}
    \vspace{-2mm}
    \resizebox{1.0\columnwidth}{!}{
\begin{tabular}{cccllll}
 \Xhline{1.0pt}
                          & \multicolumn{3}{c}{SSM}                              &                       &                           &                            \\ \cline{2-4}
\multirow{-2}{*}{Methods} & DConv                     & SiLU                  & $\bigotimes$ & \multirow{-2}{*}{FFN} & \multirow{-2}{*}{PSNR$\uparrow$ (dB)}    & \multirow{-2}{*}{SSIM$\uparrow$ (dB)}     \\ \hline
RSDhamba-V1               & {\ding{51}}                        & {\ding{51}}                     & {\ding{51}}        & \multicolumn{1}{c}{\ding{55}}                     & \multicolumn{1}{c}{25.95} & \multicolumn{1}{c}{0.9262} \\
RSDhamba-V2               & {\ding{55}}     & {\ding{51}}                     & {\ding{51}} & \multicolumn{1}{c}{{\ding{51}}}                     & \multicolumn{1}{c}{26.44} & \multicolumn{1}{c}{0.9273} \\
RSDhamba-V3               & {\ding{51}}    & {\ding{55}} & {\ding{51}} & \multicolumn{1}{c}{{\ding{51}}}                     & \multicolumn{1}{c}{26.34}                     & \multicolumn{1}{c}{0.9282}                     \\
RSDhamba-V4               & {\ding{51}}    & {\ding{51}} &  {\ding{55}} & \multicolumn{1}{c}{\ding{51}}                    & \multicolumn{1}{c}{26.00}                     & \multicolumn{1}{c}{0.9262}                     \\
\rowcolor[HTML]{EBF1FF} 
{\color[HTML]{333333} RSDhamba (Ours)} &
  {\ding{51}} &
  {\color[HTML]{000000} {\ding{51}}} &
  {\ding{51}} &
 \multicolumn{1}{c}{{\ding{51}}} &
  \multicolumn{1}{c}{\cellcolor[HTML]{EBF1FF}\textbf{26.54}} &
  \multicolumn{1}{c}{\cellcolor[HTML]{EBF1FF}\textbf{0.9299}} \\ \Xhline{1.0pt}
\end{tabular}
    }
    \vspace{-5mm}
    \label{table2}
\end{table}

\begin{table}[t]\small
    \centering
    \caption{ABLATION ANALYSIS OF THE NUMBER OF DSM CHANNEL SCANS WITHIN THE MODEL.}
    \vspace{-2mm}
    \resizebox{1.0\columnwidth}{!}{
\begin{tabular}{c|ccc|c|c}
    \Xhline{1.0pt}
Scan Settings  & DSM-1 & DSM-2                    & DSM-4                    & PSNR $\uparrow$ (dB)         & SSIM $\uparrow$  (dB)        \\ \hline
RSDhamba-DSM-1 & {\ding{51}}     & -                        & -                        & 26.15          & 0.9255          \\
RSDhamba-DSM-2 & -     & {\color[HTML]{000000} {\ding{51}}} & -                        & 26.22          & 0.9277          \\
\rowcolor[HTML]{EBF1FF} 
RSDhamba (Ours) & -     & -                        & {\color[HTML]{000000} {\ding{51}}} & \textbf{26.54} & \textbf{0.9299} \\\Xhline{1.0pt}
    \end{tabular}
    }
    \vspace{-5mm}
    \label{table3}
\end{table}

\vspace{-4mm}
\subsection{Ablation Study}
We conducted ablation experiments on the thin haze dataset to investigate the effectiveness of the various components of our proposed model. These experiments include evaluating the effectiveness of each internal component in designing the SSM, and exploring the effect of different number of scanning channels on the dehazing effect.

{\flushleft\textbf{Analysis of the proposed SSM}.} To thoroughly evaluate the effectiveness of SSM in the Mamba-based remote sensing image dehazing method, we train the network by changing the internal components of our designed SSM. The corresponding results are shown in Table \ref{table2}. It can be seen that our designed SSM method has better improvement in PSNR and SSIM metrics compared to other SSM modules.

{\flushleft\textbf{Analysis of the number of scanning paths}.} We perform an analytical study on the thin haze dataset to analyze the number of scanning paths. Table \ref{table3} shows the impact of single-path, two-path, and four-path scans on the recoverability. The results show that the four-path scanning we designed usually leads to better performance.

\section{Conclusion}
In this letter, we present the first lightweight network on the mamba-based model in the field of RSID, called RSDhamba.
To thoroughly explore the spatially varying distributional information of haze, we designed a three-level encoder-decoder structure with Vision Dehamba Block (VDB) forming the backbone element of the network and integrated the SSM framework into the U-Net architecture. 
Furthermore, we propose a Direction-aware Scan Module (DSM) to better perceive the spatially varying distribution of haze from different directions and realize dynamic feature fusion in multiple directions.
Extensive benchmark experimental results show that RSDhamba has favorable performance compared with state-of-the-art methods. 
In future work, we intend to explore the potential applications of our developed model in other RS visualization domains.

\vspace{-1mm}
\bibliographystyle{IEEEtran}
\bibliography{IEEEfull}
\end{document}